\documentclass[11pt]{article}

\usepackage[final]{acl-style-files-master/acl}
\usepackage{afterpage}
\usepackage{times}
\usepackage{latexsym}

\usepackage[T1]{fontenc}

\usepackage[utf8]{inputenc}

\usepackage{inconsolata}

\usepackage{graphicx}
\usepackage{stfloats}
\usepackage{tikz}

\usepackage{array} 
\usepackage{booktabs}
\usepackage{multirow} 
\usepackage{xcolor} 
\usepackage{tcolorbox} 
\tcbuselibrary{breakable,skins} 
\usepackage{setspace} 
\usepackage{pifont} 
\usepackage{float}

\definecolor{userpromptframecolor}{gray}{0.50} 
\definecolor{userpromptbackcolor}{gray}{0.80}  
\definecolor{userprompttextcolor}{HTML}{000000} 

\definecolor{assistantpromptframecolor}{HTML}{A8D8FF} 
\definecolor{assistantpromptbackcolor}{HTML}{E6F3FF}  
\definecolor{assistantprompttextcolor}{HTML}{003366}  

\definecolor{redcrosscolor}{HTML}{CC0000}   
\definecolor{greencheckcolor}{HTML}{008000} 

\definecolor{interval0back}{HTML}{FFFFFF}   
\definecolor{interval25back}{HTML}{F2F9FF}  
\definecolor{interval50back}{HTML}{E6F3FF}  
\definecolor{interval75back}{HTML}{CDE2F7}  
\definecolor{interval100back}{HTML}{B4D2F0} 
\definecolor{intervalframecolor}{HTML}{8BB8D9} 

\newcommand{\figureinput}[1]{%
  \begingroup%
  \setlength{\fboxsep}{1pt}
  \fcolorbox{userpromptframecolor}{userpromptbackcolor}{%
    \color{userprompttextcolor}\ttfamily\scriptsize #1%
  }%
  \endgroup%
}

\newcommand{\inlineinput}[1]{%
  \begingroup%
  \setlength{\fboxsep}{1pt}
  \fcolorbox{userpromptframecolor}{userpromptbackcolor}{%
    \color{userprompttextcolor}\ttfamily\scriptsize #1%
  }%
  \endgroup%
}

\newcommand{\inlineoutput}[1]{%
  \begingroup%
  \setlength{\fboxsep}{1pt}
  \fcolorbox{assistantpromptframecolor}{assistantpromptbackcolor}{%
    \color{assistantprompttextcolor}\ttfamily\scriptsize #1%
  }%
  \endgroup%
}

\newcommand{\markfont}{\fontsize{7pt}{9pt}\selectfont}

\newcommand{\correctmark}{{\color{greencheckcolor}\markfont\ding{51}}} 
\newcommand{\incorrectmark}{{\color{redcrosscolor}\markfont\ding{55}}} 

\tcbset{
    chatpromptbase/.style={
        fontupper=\setstretch{1.2}\ttfamily\scriptsize, 
        boxrule=0.5pt,          
        boxsep=1pt,             
        left=3pt, right=3pt, top=2pt, bottom=2pt, 
        before skip=0pt,        
        after skip=0pt,         
        before upper={\parindent0pt}, 
        arc=0pt
    }
}

\newtcolorbox{usergrayprompt}{
    chatpromptbase, 
    colback=userpromptbackcolor,    
    coltext=userprompttextcolor,    
    colframe=userpromptframecolor,  
}

\newtcolorbox{assistantblueprompt}{
    chatpromptbase, 
    colback=assistantpromptbackcolor,   
    coltext=assistantprompttextcolor,   
    colframe=assistantpromptframecolor, 
}

\newtcolorbox{intervalboxouter}[3]{%
    enhanced,
    colback=#2,
    colframe=intervalframecolor,
    coltext=assistantprompttextcolor,
    fontupper=\setstretch{1.2}\ttfamily\scriptsize,
    boxrule=0.5pt,
    boxsep=0pt,
    left=0pt, right=#1, top=0pt, bottom=0pt,
    before skip=0pt, after skip=0pt,
    before upper={\parindent0pt},
    arc=0pt,
    overlay={\node[anchor=south east, font=\sffamily\tiny\bfseries,
                   text=assistantprompttextcolor, fill=#2,
                   inner sep=1pt]
                   at ([xshift=-2pt, yshift=2pt]frame.south east) {#3};}
}
\newtcolorbox{intervalbox}[3]{%
    enhanced,
    colback=#2,
    colframe=intervalframecolor,
    coltext=assistantprompttextcolor,
    fontupper=\setstretch{1.2}\ttfamily\scriptsize,
    boxrule=0.5pt,
    toprule=0pt, leftrule=0pt,  
    boxsep=0pt,
    left=0pt, right=#1, top=0pt, bottom=0pt,
    before skip=0pt, after skip=0pt,
    before upper={\parindent0pt},
    arc=0pt,
    overlay={\node[anchor=south east, font=\sffamily\tiny\bfseries,
                   text=assistantprompttextcolor, fill=#2,
                   inner sep=1pt]
                   at ([xshift=-2pt, yshift=2pt]frame.south east) {#3};}
}
\newcommand{\ptext}[1]{\par\vspace{2pt}\noindent\hspace{3pt}\parbox[t]{\dimexpr\linewidth-6pt\relax}{\raggedright #1}\vspace{2pt}}

\newcommand{\imageoverlay}[2]{%
    \begin{tikzpicture}
        \node[anchor=south west, inner sep=0pt] (image) {\includegraphics[width=0.15\linewidth]{#2}};
        \node[anchor=south east, xshift=-2pt, yshift=2pt, 
              fill=white, fill opacity=0.75, 
              text opacity=1, 
              rounded corners=1.5pt, inner sep=1pt, 
              font=\sffamily\scriptsize\bfseries] 
              at (image.south east) {#1};
    \end{tikzpicture}%
}

\title{Prefill-Guided Thinking for zero-shot detection of AI-generated images}

\author{
  Zoher Kachwala\thanks{Corresponding author: \texttt{zkachwal@iu.edu}} \quad
  Danishjeet Singh \quad
  Danielle Yang \quad
  Filippo Menczer \\
  Observatory on Social Media \\
  Indiana University, Bloomington, USA
}

\begin{document}

\maketitle

\begin{abstract}
Traditional supervised methods for detecting AI-generated images depend on large, curated datasets for training and fail to generalize to novel, out-of-domain image generators. 
As an alternative, we explore pre-trained Vision-Language Models (VLMs) for zero-shot detection of AI-generated images. 
We evaluate VLM performance on three diverse benchmarks encompassing synthetic images of human faces, objects, and animals produced by 16 different state-of-the-art image generators. 
While off-the-shelf VLMs perform poorly on these datasets, we find that prefilling responses effectively guides their reasoning --- a method we call Prefill-Guided Thinking (\textbf{PGT}). 
In particular, prefilling a VLM response with the phrase ``\textit{Let's examine the style and the synthesis artifacts}'' improves the Macro F1 scores of three widely used open-source VLMs by up to 24\%.
We analyze this improvement in detection by tracking answer confidence during response generation.
For some models, prefills counteract early overconfidence --- akin to mitigating the Dunning-Kruger effect --- leading to better detection performance.
\end{abstract}

\begin{figure*}[b]
\centering

\setlength{\tabcolsep}{2pt}
\resizebox{0.61\linewidth}{!}{
\begin{tabular}{ccccc}
\imageoverlay{1}{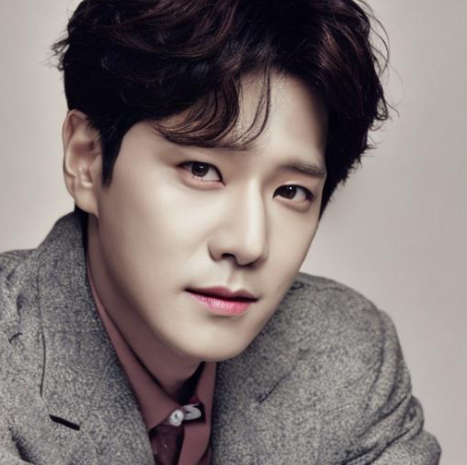} &
\imageoverlay{2}{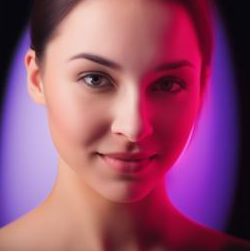} &
\imageoverlay{3}{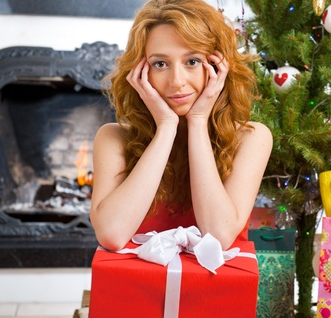} &
\imageoverlay{4}{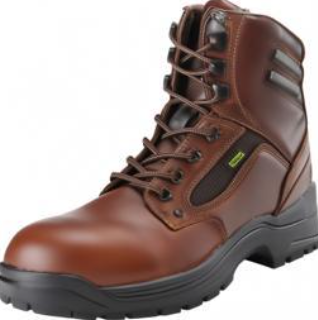} &
\imageoverlay{5}{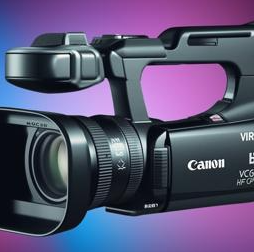} \\

\imageoverlay{6}{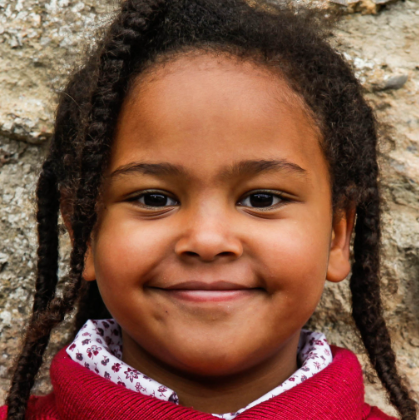} &
\imageoverlay{7}{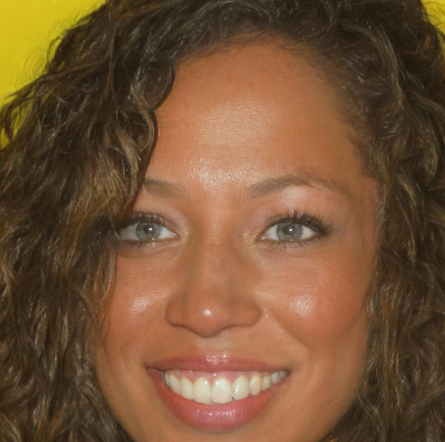} &
\imageoverlay{8}{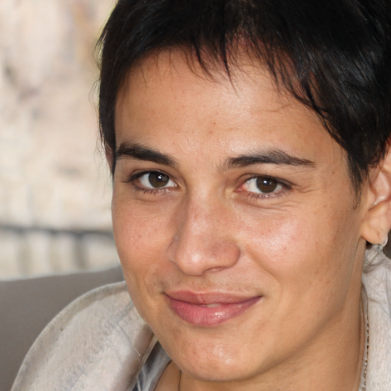} &
\imageoverlay{9}{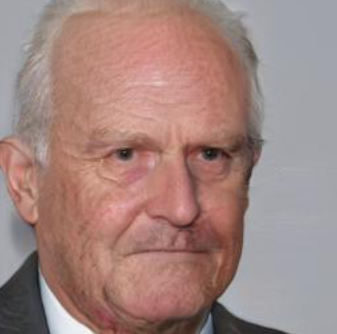} &
\imageoverlay{10}{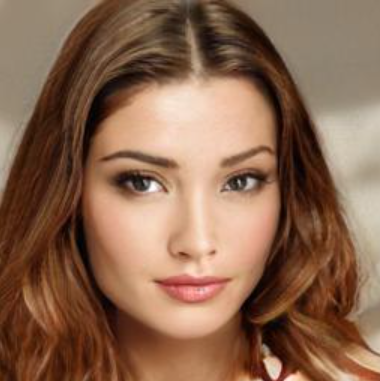} \\

\imageoverlay{11}{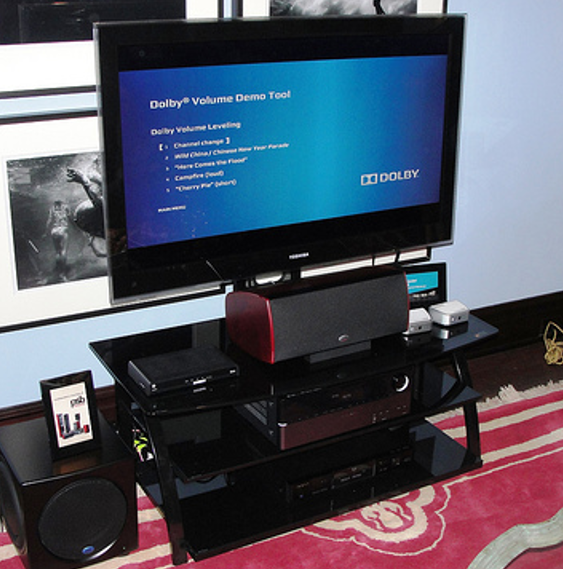} &
\imageoverlay{12}{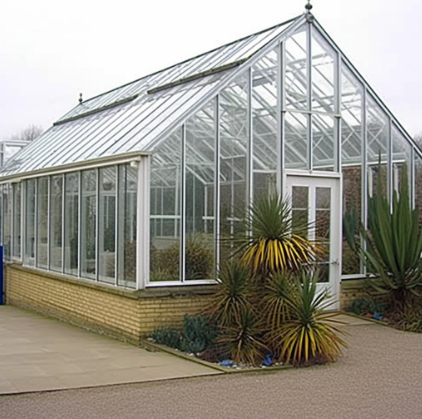} &
\imageoverlay{13}{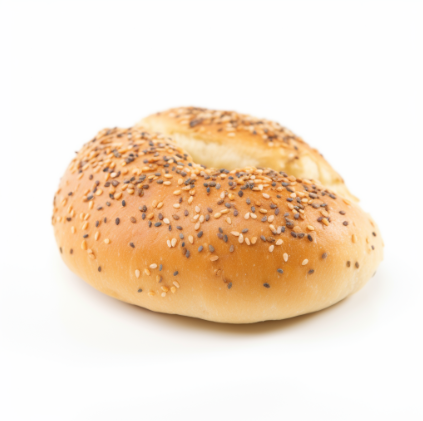} &
\imageoverlay{14}{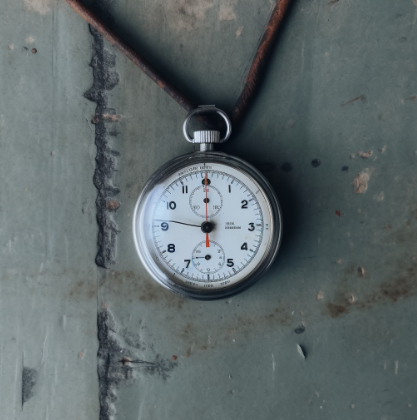} &
\imageoverlay{15}{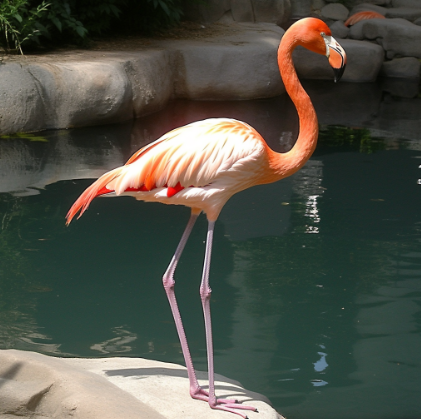}
\end{tabular}
}

\vspace{1em}

\begin{minipage}{\linewidth}
\begin{minipage}[t]{0.495\linewidth}
\begin{usergrayprompt}
User: [Image] Is this image real or AI‑generated?
\end{usergrayprompt}
\par\vspace{0.5mm}
\begin{assistantblueprompt}
Assistant: \figureinput{Let's think step by step}
\end{assistantblueprompt}
\end{minipage}%
\hfill
\begin{minipage}[t]{0.495\linewidth}
\begin{usergrayprompt}
User: [Image] Is this image real or AI‑generated?
\end{usergrayprompt}
\par\vspace{0.5mm}
\begin{assistantblueprompt}
Assistant: \figureinput{Let's examine the style and synthesis artifacts}
\end{assistantblueprompt}
\end{minipage}
\end{minipage}

\caption{Top: Sample images from D3 (top row), DF40 (middle row), and GenImage (bottom row). Can you identify the real ones? (Answer in footnote.) Bottom: Guiding model responses via prefilling: chain-of-thought (left) vs task-aligned (right).}
\label{fig:my_image_grid}
\end{figure*}

\section{Introduction}

Rapid advancements in image generation have led to a surge in synthetic images (deepfakes) \citep{goodfellow_generative_2014, ho_denoising_2020, rombach_high-resolution_2022}. 
Improved techniques now enable the easier and cheaper production of high-quality visuals \citep{rombach_high-resolution_2022, li_blip-2_2023, dhariwal_diffusion_2021}. 
These techniques benefit creative applications but also enable malicious actors to produce convincing forgeries (e.g., face swaps, synthetic photos) that are nearly indistinguishable from real ones \citep{li_advancing_2020, lugmayr_repaint_2022, lu_seeing_2023}. 
Such forgeries facilitate impersonation, copyright infringement, and disinformation, highlighting the urgent need for robust detection methods to maintain visual trust \citep{yang_characteristics_2024, somepalli_diffusion_2023, diresta_how_2024, chandra_analyzing_2024}.

Existing solutions remain limited. 
Watermarking and metadata approaches are often easily bypassed and require widespread adoption \citep{zhao_invisible_2024}. 
Supervised methods, particularly feature-based ones, struggle to generalize to new generators \citep{li_freqblender_2024}. 

To keep pace with these rapid developments, we investigate the use of pre-trained Vision-Language Models (VLMs), which have demonstrated strong generalization across diverse tasks \citep{radford_learning_2021, li_blip_2022, li_blip-2_2023} and whose performance can be further enhanced through prompting \citep{yang_fine-grained_2023, du_ipo_2024}. 

We find that off-the-shelf VLMs perform poorly on the zero-shot detection of AI-generated images. 
However, prefilling a model response can guide more focused reasoning and significantly improve detection without any fine-tuning. 
We call this approach Prefill-Guided Thinking (\textbf{PGT}). 
In particular, prefilling with \textit{``Let's examine the style and the synthesis artifacts''} directs the VLM toward forensic cues. 

We evaluate \textbf{PGT} on three diverse datasets spanning human faces, objects, and animals. 
These images have been generated by 16 state-of-the-art image models that produce photorealistic outputs lacking visible synthesis artifacts (Fig.~\ref{fig:my_image_grid}\footnote{\rotatebox{180}{Images 3, 10, and 11 are the only real ones in Fig.~\ref{fig:my_image_grid}.}}).
Despite this, our approach boosts Macro F1 scores by up to 24\% across three open-source VLMs, generalizing across image categories and modern generators.

We explore why prefilling helps by tracking answer confidence and Macro F1 as responses progress.
Baseline queries trigger immediate high confidence despite poor detection --- the model commits to an answer before examining the image.
For Qwen and LLaVA on most datasets, prefills counteract this overconfidence more effectively than prompts, inducing uncertainty that leads to gradually improving performance as the model's response progresses. 

\section{Background}

Supervised detection methods for AI-generated images typically fall into three categories: artifact-based, frequency-domain, and spatial-domain approaches.

\textit{Artifact-based methods} use Convolutional Neural Networks (CNNs) or Vision Transformers (ViTs) to detect subtle cues such as unnatural textures or edge inconsistencies \citep{afchar_mesonet_2018, rossler_faceforensics_2019, chollet_xception_2017}.  
As generative models produce more realistic outputs, texture and edge artifacts become harder to detect.  
Models trained on fixed artifacts often overfit to specific generators, leading to poor generalization \citep{wang_cnn-generated_2020, cozzolino_raising_2024, ojha_towards_2023}.

\textit{Frequency-domain techniques} analyze spectral representations using techniques like Fast Fourier Transform or Discrete Cosine Transform \citep{li_freqblender_2024, dzanic_fourier_2020, frank_leveraging_2020, tan_frequency-aware_2024}.  
These methods detected early Generative Adversarial Networks (GANs) effectively, but diffusion models produce frequency patterns that fixed-frequency detectors miss \citep{cozzolino_raising_2024, baraldi_contrasting_2024, zheng_breaking_2024}.

\textit{Spatial-domain approaches} examine raw pixel patterns to detect structural or textural inconsistencies \citep{zheng_breaking_2024, zhou_learning_2018}.  
While useful for earlier synthetic images, they often fail on photorealistic outputs from diffusion models and are sensitive to post-processing such as compression and resizing \citep{rossler_faceforensics_2019, bammey_synthbuster_2024}.

To improve generalization, recent work uses larger and more diverse training datasets \citep{wang_cnn-generated_2020, yan_deepfakebench_2023, ye_loki_2024, zhu_genimage_2023, yan_df40_2024, zhang_perceptual_2023, shao_detecting_2023}, as well as architectural modifications that target universal artifacts (e.g., upsampling patterns) \citep{zheng_breaking_2024, tan_rethinking_2024}.  
Approaches like fixed-feature backbones and DiffusionFake have also improved robustness to novel generators \citep{ojha_towards_2023, sun_diffusionfake_2024}.

Vision-Language Models (VLMs) are trained on large-scale image-text datasets and demonstrate strong zero-shot generalization across tasks such as classification, captioning, and visual question answering \citep{li_blip-2_2023, radford_learning_2021, li_blip_2022}.  
Prompt-based learning has emerged as a lightweight method to adapt language models using natural language instructions \citep{brown_language_2020}.  
Chain-of-thought, which encourages step-by-step reasoning, has been effective in multimodal settings \citep{wei_chain--thought_2022, kojima_large_2022, gu_systematic_2023}.

Given the success of chain-of-thought prompting, we investigate whether prefilling VLM responses can enable zero-shot detection of AI-generated images.
If successful, this would avoid the need for generator-specific training data. 

\section{Methods}
\label{sec:methods}

\begin{figure*} 
  \centering 
  \begin{minipage}[c]{0.383\textwidth} 
      \centering 
      \includegraphics[width=\linewidth, keepaspectratio]{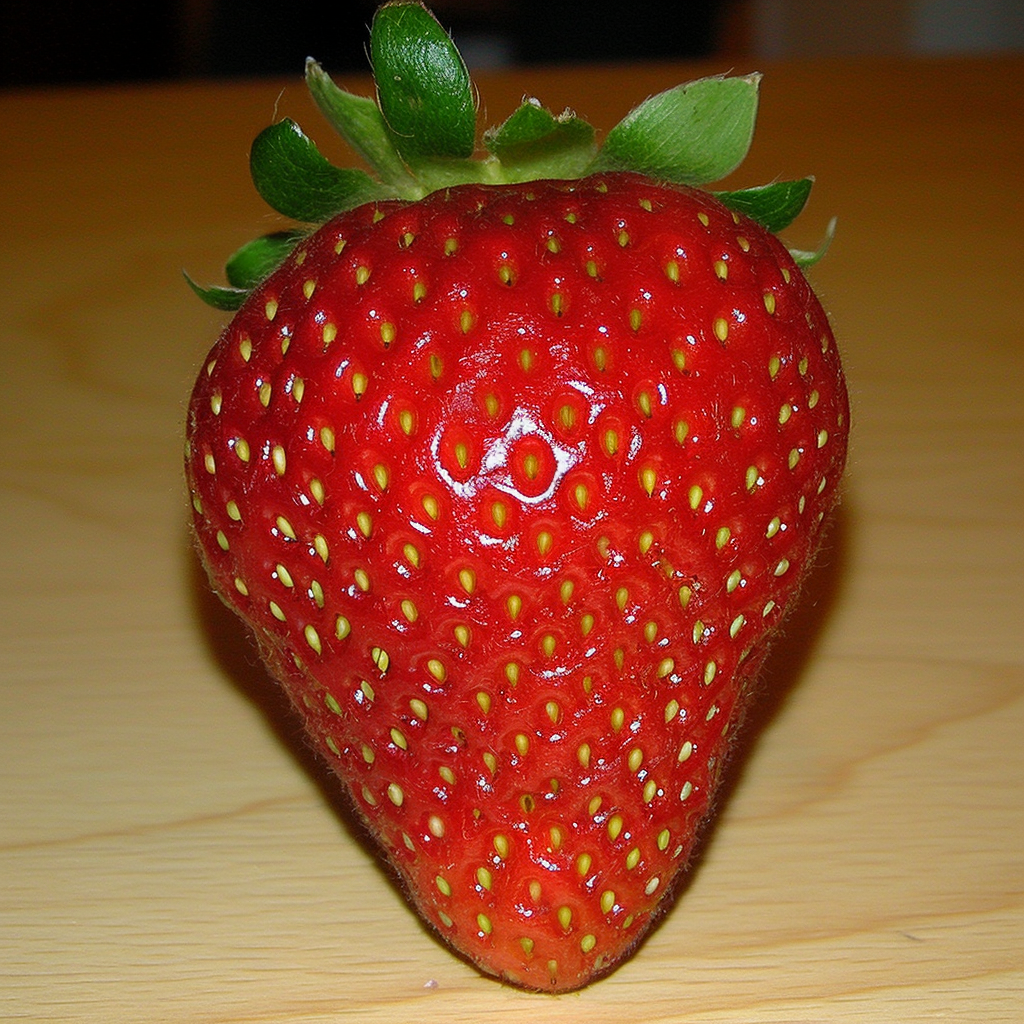} 
  \end{minipage}
  \begin{minipage}[c]{0.57\textwidth} 
      
      \noindent 
      \begin{minipage}[t]{0.05\linewidth} 
          \raisebox{-8pt}{ \ttfamily\scriptsize a)}
      \end{minipage}%
      \begin{minipage}[t]{.94\linewidth} 
          \begin{usergrayprompt}
              User: [Image] Is this image real or AI‑generated?
          \end{usergrayprompt}
          \par\vspace{0.5mm} 
          \begin{assistantblueprompt}
              Assistant: The image appears to be real. The details, such as the texture of the strawberry's surface...\textit{\scriptsize (continues)}
              \par\noindent{\figureinput{Final Answer(real/ai‑generated):}} real. \incorrectmark 
          \end{assistantblueprompt}
      \end{minipage}
      \vspace{4pt} 

      \noindent
      \begin{minipage}[t]{0.05\linewidth} 
          \raisebox{-8pt}{ \ttfamily\scriptsize b)}
      \end{minipage}%
      \begin{minipage}[t]{.94\linewidth} 
          \begin{usergrayprompt}
              User: [Image] Is this image real or AI‑generated?
          \end{usergrayprompt}
          \par\vspace{0.5mm} 
          \begin{assistantblueprompt}
              Assistant: \figureinput{Let's think step by step}
              \par1. Texture and Detail: The strawberry appears to have a very realistic texture...\textit{\scriptsize (continues)}
              \par\noindent{\figureinput{Final Answer(real/ai‑generated):}} real. \incorrectmark 
          \end{assistantblueprompt}
      \end{minipage}
      \vspace{4pt} 

      \noindent
      \begin{minipage}[t]{0.05\linewidth} 
          \raisebox{-8pt}{ \ttfamily\scriptsize c)}
      \end{minipage}%
      \begin{minipage}[t]{0.94\linewidth} 
          \begin{usergrayprompt}
              User: [Image] Is this image real or AI‑generated?
          \end{usergrayprompt}
          \par\vspace{0.5mm} 
          \begin{assistantblueprompt}
              Assistant: \figureinput{Let's examine the style and the synthesis artifacts}
              \par1. Texture and Detail: The strawberry appears to have a very smooth and uniform texture...\textit{\scriptsize (continues)}
              \par\noindent{\figureinput{Final Answer(real/ai‑generated):}} ai‑generated. \correctmark 
          \end{assistantblueprompt}
      \end{minipage}
  \end{minipage}
  \caption{\textbf{PGT} applied to a Midjourney-generated image of a strawberry, using Qwen2.5-7B as the VLM.
Input text in \inlineinput{grey}; model responses in \inlineoutput{blue}.
(a)~A direct query produces the incorrect answer \inlineoutput{real}.
(b)~The chain-of-thought prefill \inlineinput{Let's think step by step} elicits step-by-step analysis, but the classification remains incorrect.
(c)~Our \textbf{S2} prefill \inlineinput{Let's examine the style and the synthesis artifacts} yields the correct classification: \inlineoutput{ai‑generated}.
Full reasoning traces appear in the Appendix.}
  \label{fig:strawberry}
\end{figure*}

We frame the detection of AI-generated images as a binary classification task: given an image, the goal is to determine whether it is \textit{real} or \textit{AI-generated}. 
To evaluate overall performance, we use the Macro F1 score, which is robust to class imbalance.
To analyze performance across different generators, we additionally report per-generator recall within the \textit{AI-generated} class.

\paragraph{Prefill-Guided Thinking.}
Vision-Language Models typically consist of a \textit{system} field for general instructions, a \textit{user} field for inputs, and an \textit{assistant} field for model-generated responses.    
A \textit{prefill} is a phrase inserted into the \textit{assistant} field, intended to guide the model's response.

In our experiment, an image is presented to a VLM along with a question in the user field: \inlineinput{User: [Image] Is this image real or AI-generated?}.
The VLM generates free-form reasoning in the assistant field (e.g., \inlineoutput{Assistant: This image appears to be...}) (Fig.~\ref{fig:strawberry}a).
To obtain a clean label, we query the model a second time by inserting a phrase into the assistant field after its first response: \inlineinput{Final Answer(real/ai-generated):}.
This constrains the model to select between the two labels presented in the prompt.
We extract predictions using keyword matching for \textit{real} and \textit{ai-generated} (or its synonyms: \textit{artificial}, \textit{fake}, \textit{generated}), defaulting to real when both appear.
This is our \textit{baseline} method. 

The chain-of-thought (\textit{CoT}) variant encourages step-by-step reasoning.
It uses the same initial question in the user field and inserts the phrase \textit{``Let's think step by step''} into the assistant field \citep{kojima_large_2022}.
The model's prefilled responses then start with: \inlineoutput{Assistant: \inlineinput{Let's think step by step}} (Fig.~\ref{fig:strawberry}b).  
The final clean label is again elicited using the same follow-up answer phrase in the assistant field.

Building on prior work that highlights the importance of synthesis artifacts in detection \citep{jia_can_2024}, we introduce \textbf{S2}, a task-aligned prefill.
We insert the phrase \textit{``Let's examine the style and the synthesis artifacts''} into the assistant's field (Fig.~\ref{fig:strawberry}c), creating a prefilled response that starts with: \inlineoutput{Assistant: \inlineinput{Let's examine the style and the synthesis artifacts}}.
This framing directs the model toward perceptual cues --- stylistic inconsistencies or generation artifacts --- rather than semantic features. 
The final classification label is obtained in the same way as in the baseline. 

We also evaluate prefill variations and compare inserting phrases into the user field (prompting) or system field. 

\paragraph{Data.}
We conduct experiments using three diverse benchmarks spanning multiple visual domains and 16 state-of-the-art image generators.
Together, these datasets encompass human faces, natural objects, animals, urban scenes, artwork, and abstract imagery, with synthetic images produced by both GAN and diffusion-based architectures.

\textit{D3} is a benchmark dataset introduced as part of the Contrastive Deepfake Embeddings framework \citep{baraldi_contrasting_2024}. 
Unlike many generative datasets focused on faces or curated categories, D3 comprises real images collected from the web, covering a wide range of domains, including objects, urban scenes, artwork, animals, abstract visuals, and human figures.
Synthetic counterparts were generated using four models: DeepFloyd IF, Stable Diffusion v1.4 and v2.1 \citep{rombach_high-resolution_2022}, and Stable Diffusion XL \citep{podell_sdxl_2023}.
We randomly sampled 2,000 sets of five images (one real and four generated).
After filtering for copyright restrictions and broken links, the final dataset contains 8,420 images (1,684 real and 6,736 generated).
We use 80\% of this data (1,344 real and 5,392 generated) for our main evaluation and reserve the remaining 20\% (344 real and 1,344 generated) for additional experiments.
We refer to the main evaluation set as D3, and the smaller subset as D3~(2k).

\textit{DF40} is a facial image dataset containing content generated by 40 deepfake techniques across four categories: face swapping, face reenactment, full-face synthesis, and facial editing \citep{yan_df40_2024}. 
It includes outputs from state-of-the-art models such as Collaborative Diffusion \citep{huang_collaborative_2023}, Midjourney, StyleCLIP \citep{patashnik_styleclip_2021}, StarGAN v1 and v2 \citep{choi_stargan_2018, choi_stargan_2020}, and WhichFaceIsReal. 
The dataset spans variations in age, gender, ethnicity, and facial pose. 
We randomly sample 10,000 images (3,929 real and 6,071 generated) for the main evaluation, and an additional 2,000 images (794 real and 1,206 generated) for extended experiments. We refer to the main evaluation set as DF40, and the smaller subset as DF40~(2k).

\begin{figure*}[!t]  
    \centering
    \includegraphics[width=\textwidth]{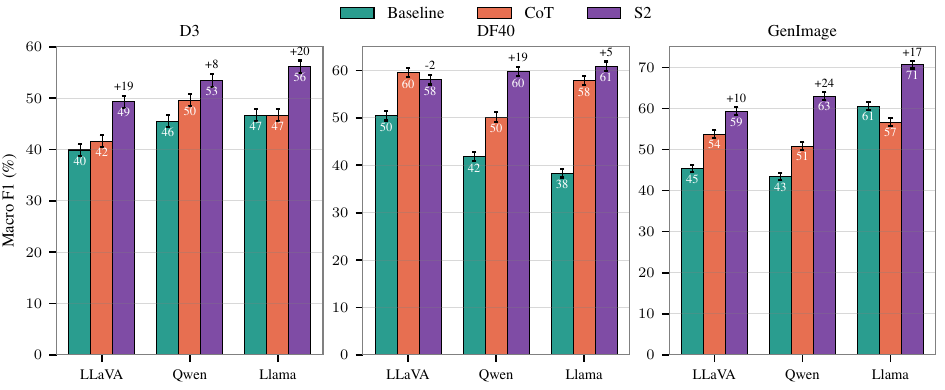}
      \caption{
      Detection Macro F1 across models, datasets, and \textbf{PGT} variations.
      Bars show relative improvement of \textbf{S2} over the next best method, with 95\% confidence intervals from 10k bootstrap iterations.
      }
    \label{fig:macro_f1_bars}
\end{figure*}

\begin{figure*}[!t]
\centering
\includegraphics[width=\textwidth]{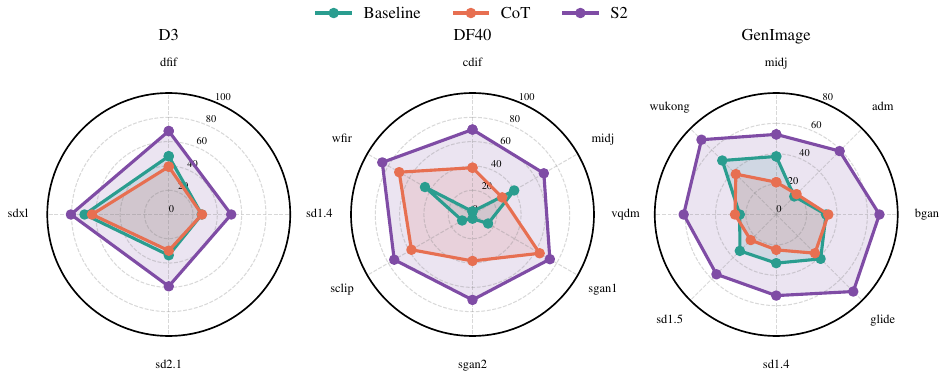}
\caption{
Detection recall (\%) for Llama on each dataset, broken down by generator. Similar figures for Qwen and LLaVA are in the Appendix (Figs.~\ref{fig:recall_radar_qwen}, \ref{fig:recall_radar_llava}).
}
\label{fig:recall_radar_llama}
\end{figure*}

\textit{GenImage} is built using all real images from ImageNet \citep{deng_imagenet_2009}, covering diverse object categories such as animals, tools, vehicles, and furniture.
Each real image is paired with a synthetic counterpart generated by one of eight models: ADM \citep{dhariwal_diffusion_2021}, BigGAN \citep{brock_large_2018}, GLIDE \citep{nichol_glide_2022}, Midjourney, VQDM \citep{gu_vector_2022}, Stable Diffusion v1.4 and v1.5 \citep{rombach_high-resolution_2022}, and Wukong.
We use a balanced evaluation sample of 10,000 images (5,000 real and 5,000 generated) for the main experiments and 2,000 images (1,000 real and 1,000 generated) for extended analyses.
We refer to the main evaluation set as GenImage, and the smaller subset as GenImage~(2k).

Collectively, our evaluation spans 28,420 images across diverse visual content and 16 generators.
Modern diffusion models in these benchmarks --- Stable Diffusion XL, GLIDE, Midjourney --- produce photorealistic images that lack traditional synthesis artifacts (Fig.~\ref{fig:my_image_grid}).

\paragraph{Models.}

\begin{table*}[t]
    \centering
    \begin{tabular}{lllll}
    \toprule
    \textbf{Phrase} & \textbf{Type} & \textbf{D3~(2k)} & \textbf{DF40~(2k)} & \textbf{GenImage~(2k)} \\
    \midrule
    Baseline & --- & 46.7 & 42.3 & 44.5 \\
    \midrule
    \multirow{3}{*}{CoT} & Prefill & 49.8 & 48.8 & 53.3 \\
     & Pseudo-Prefill & 48.9 {\scriptsize (-0.9)} & 44.1 {\scriptsize (-4.7)} & 48.8 {\scriptsize (-4.5)} \\
     & Prompt & 45.5 {\scriptsize (-4.3)} & 46.7 {\scriptsize (-2.1)} & 50.4 {\scriptsize (-2.9)} \\
    \cmidrule(lr){1-5}
    \multirow{3}{*}{S2} & Prefill & 53.2 & 61.2 & 64.8 \\
     & Pseudo-Prefill & 43.4 {\scriptsize (-9.8)} & 45.2 {\scriptsize (-16.0)} & 47.3 {\scriptsize (-17.5)} \\
     & Prompt & 49.7 {\scriptsize (-3.5)} & 48.8 {\scriptsize (-12.4)} & 50.2 {\scriptsize (-14.6)} \\
    \bottomrule
    \end{tabular}
\caption{Macro F1 scores (\%) across datasets and guidance modes for Qwen. Values in parentheses show differences from the corresponding prefill. Tables for LLaVA and Llama are in the Appendix.}
\label{tab:macro_f1_modes_qwen25_vl_7b}
\end{table*}

We conduct experiments with three Vision-Language Models: \textit{LLaVA-OneVision}\footnote{\url{https://huggingface.co/collections/llava-hf/llava-onevision-66bb1e9ce8856e210a7ed1fe}}, \textit{Qwen2.5-VL}\footnote{\url{https://huggingface.co/collections/Qwen/qwen25-vl-6795ffac22b334a837c0f9a5}}, and \textit{Llama-3.2-Vision}\footnote{\url{https://huggingface.co/collections/meta-llama/llama-32-66f448ffc8c32f949b04c8cf}}.
We use instruction-tuned variants in evaluation mode and use \texttt{seed=0} in the vLLM package \citep{kwon_efficient_2023} to ensure reproducibility.
Token generation is capped at 512 tokens (\texttt{max\_tokens=512}) with \texttt{temperature=0.0} for deterministic outputs. 
A single run on a NVIDIA A100 80GB for 10,000 images takes approximately 3 hours.

\begin{table*}[t]
    \centering
    \begin{tabular}{llllll}
    \toprule
    \textbf{Phrase} & \textbf{Prefill} & \textbf{D3~(2k)} & \textbf{DF40~(2k)} & \textbf{GenImage~(2k)} \\
    \midrule
    Baseline & --- & 46.7 & 42.3 & 44.5 \\
    \midrule
    CoT & \figureinput{Let's think step by step} & 49.8 & 48.8 & 53.3 \\
    \midrule
    S2 & \figureinput{Let's examine the style and the synthesis artifacts} & 53.2 & 61.2 & 64.8 \\
    \midrule
    \multirow{5}{*}{Variants} & \figureinput{Let's observe the style and the synthesis artifacts} & 55.6 {\scriptsize (+2.4)} & 62.3 {\scriptsize (+1.1)} & 64.6 {\scriptsize (-0.2)} \\
     & \figureinput{Let's examine the synthesis artifacts} & 54.6 {\scriptsize (+1.4)} & 60.9 {\scriptsize (-0.3)} & 64.0 {\scriptsize (-0.8)} \\
     & \figureinput{Let's examine the style} & 48.5 {\scriptsize (-4.7)} & 55.7 {\scriptsize (-5.5)} & 54.6 {\scriptsize (-10.2)} \\
     & \figureinput{Let's examine the details} & 49.2 {\scriptsize (-4.0)} & 50.9 {\scriptsize (-10.3)} & 51.2 {\scriptsize (-13.6)} \\
     & \figureinput{Let's examine the flaws} & 58.3 {\scriptsize (+5.1)} & 40.0 {\scriptsize (-21.2)} & 54.1 {\scriptsize (-10.7)} \\
    \bottomrule
    \end{tabular}
\caption{Macro F1 scores (\%) of different prefill phrasings for Qwen. Variants are annotated to show absolute differences compared to \textbf{S2}.}
\label{tab:prefill_sensitivity}
\end{table*}

\textit{LLaVA-OneVision} is a multimodal instruction-following model; we use the 7B chat variant.

\textit{Qwen2.5-VL} uses a dynamic-resolution Vision Transformer \citep{bai_qwen25-vl_2025}; we use the 7B-Instruct variant.

\textit{Llama-3.2-Vision} pairs a vision adapter with a pre-trained Llama 3.1 language model; we use the 11B-Instruct variant.

\section{Results}

\paragraph{Detection Performance.}
With the exception of LLaVA on DF40, \textbf{S2} consistently outperforms \textit{CoT} and the baseline across all three benchmarks and models (Fig.~\ref{fig:macro_f1_bars}), achieving up to 24\% relative improvement in Macro F1.
Across all 16 generators, \textbf{S2} reliably boosts detection recall.
The gains in recall range from 15\% for DF40's sgan1 to 200\% for GenImage's adm compared to the next best method for Llama (Fig.~\ref{fig:recall_radar_llama}).
Similar trends hold for the other two models (Figs.~\ref{fig:recall_radar_qwen}, \ref{fig:recall_radar_llava} in Appendix).
Strong recall on Stable Diffusion XL, GLIDE, and Midjourney shows that \textbf{S2} detects photorealistic outputs without traditional synthesis artifacts.
The task-aligned prefill consistently outperforms chain-of-thought.

\paragraph{Guidance Mode Comparison.} 
We also compare prefill-guided thinking against simple user prompts that combine \inlineinput{User: [Image] Is this image real or AI-generated?} with \inlineinput{Please think step by step.} (\textit{CoT}) and \inlineinput{Please examine the style and synthesis artifacts.} (\textbf{S2}).
We also test pseudo-prefills, which use the system field to instruct the model: \inlineinput{System: Please start your responses with} followed by \inlineinput{``Let's think step by step''} or \inlineinput{``Let's examine the style and synthesis artifacts''}.

As shown in Table~\ref{tab:macro_f1_modes_qwen25_vl_7b}, user prompts and pseudo-prefills do not work as well as prefill-guided thinking for \textit{CoT} and \textbf{S2} using Qwen. Similar trends are observed using LLaVA, whereas user prompts do work better with Llama (Tables~\ref{tab:macro_f1_modes_llava_onevision_7b}, \ref{tab:macro_f1_modes_llama32_vision_11b} in Appendix).

\paragraph{Prefill Phrasing.}
We examine sensitivity to prefill phrasing. 
As shown in Table~\ref{tab:prefill_sensitivity}, Qwen's detection performance varies depending on how the prefill is worded.
Guiding the model to consider both style and synthesis artifacts using alternative phrasing does not reduce performance relative to our \textbf{S2} prefill.
In contrast, guiding the model to focus on only style or only synthesis artifacts leads to a drop in detection.
Guidance around details and flaws also proves less effective than \textbf{S2}.
While phrasing affects results, any prefill guidance consistently improves performance over baseline.

\section{Interpretability}

To explore how different guidance modes affect model predictions, we track answer confidence and detection performance at incremental steps during response generation.

\paragraph{Method.}
We conduct this analysis on the three smaller evaluation subsets --- D3~(2k), DF40~(2k), GenImage~(2k) --- across all three models (Qwen, LLaVA, Llama) and three guidance modes (baseline, prefill, prompt).
To compare confidence and performance at fixed points, we use sentence counts to divide each response into five partial-response intervals: 0\%, 25\%, 50\%, 75\%, and 100\% of sentences, as illustrated in Fig.~\ref{fig:interval_example}.
For very short responses (fewer than 4 sentences), some partial responses may be empty. 
At each partial-response interval, we construct the input by concatenating the original query, the partial response, and the phrase \inlineinput{Final Answer(real/ai-generated): This image is} to elicit an answer. 
We generate a short completion (\texttt{temperature=0.0}, \texttt{max\_tokens=20}) while tracking token-level log probabilities.
Answer confidence is the mean log probability across all tokens in the generated completion.
Answer predictions (real/AI-generated) are extracted using the same keyword matching approach described in Section~\ref{sec:methods}.

For each data-model-guidance combination, we average answer confidence across all examples at each partial-response interval and compute Macro F1 from the pooled confusion matrix.
For prompt and prefill, answer confidence and Macro F1 are further averaged across the two phrases (CoT and S2).

\begin{figure*}[!t]
    \centering
    \begin{usergrayprompt}
    User: [Image] Is this image real or AI‑generated?
    \end{usergrayprompt}
    \vspace{0.5em}
    \begin{intervalboxouter}{3pt}{interval100back}{100\%}
    \begin{intervalbox}{3pt}{interval75back}{75\%}
    \begin{intervalbox}{3pt}{interval50back}{50\%}
    \begin{intervalbox}{3pt}{interval25back}{25\%}
    \begin{intervalbox}{3pt}{interval0back}{0\%}
    \ptext{Assistant: Let's examine the style and the synthesis artifacts}
    \end{intervalbox}
    \ptext{in the image:}
    \ptext{1.~\textbf{Texture and Detail}: The strawberry appears to have a very smooth and uniform texture, which is unusual for a real strawberry.}
    \end{intervalbox}
    \ptext{Real strawberries typically have a more varied texture with slight imperfections and variations in color and surface.}
    \ptext{2.~\textbf{Lighting and Reflection}: The lighting is very even and lacks natural variations. The reflections and highlights are too perfect and uniform, which is a common sign of AI-generated images.}
    \end{intervalbox}
    \ptext{3.~\textbf{Color and Contrast}: The color is vibrant and consistent, but the contrast is too high. Real strawberries have a more subtle contrast between the red and the green leaves.}
    \ptext{4.~\textbf{Overall Composition}: The strawberry is perfectly centered and the composition is very clean, which is often a characteristic of AI-generated images.}
    \end{intervalbox}
    \ptext{Real photographs tend to have more natural imperfections and less perfect symmetry.}
    \ptext{Based on these observations, it is likely that this image is AI-generated.}
    \end{intervalboxouter}
    \vspace{0.5em}
    \caption{Example \textbf{S2} prefill response at five partial-response intervals (0--100\% of sentences). Nested boxes show progressively longer completions, each probed for answer confidence.}
    \label{fig:interval_example}
  \end{figure*}
  
  \begin{figure*}[!t]
      \centering
      \includegraphics[width=\textwidth]{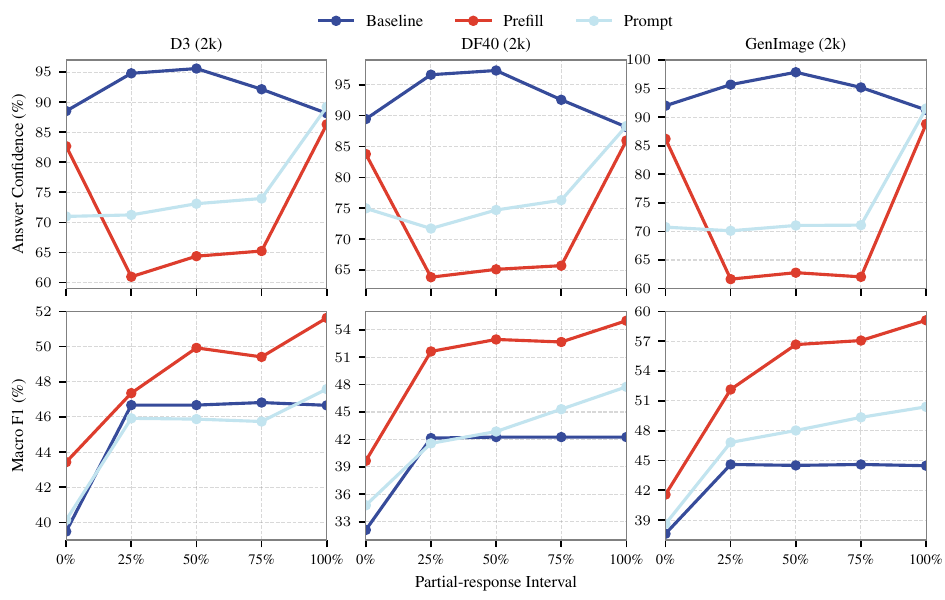}
      \caption{Evolution of answer confidence and Macro F1 across partial responses for Qwen on D3~(2k), DF40~(2k), and GenImage~(2k) using baseline, prefill, and prompt guidance modes.}
      \label{fig:interval_progression_qwen}
  \end{figure*}

\paragraph{Results.}
Models guided by prefills show markedly different reasoning patterns compared to baseline and prompt methods.
For baseline queries, Qwen exhibits high confidence (over 85\%) from the start, which remains stable throughout its response (Fig.~\ref{fig:interval_progression_qwen}).
Yet this high confidence does not translate to strong detection.
Macro F1 plateaus at 42--47\% after the first partial-response interval and remains flat.
We interpret this as early commitment: the model selects an answer before examining the image closely, then generates reasoning that supports that choice.

Prefills reverse this pattern:
Qwen's answer confidence drops sharply at 25\%, remains low through 50--75\%, then recovers at 100\%.
Crucially, detection performance improves steadily throughout this process.
Macro F1 rises from 40--45\% at the start to 52--58\% at completion --- the highest of any method.
This confidence dip --- followed by recovery only at completion --- suggests that prefills delay the model's answer selection, allowing it to accumulate image-based observations before deciding.

Prompt-guided responses maintain flat confidence throughout, with F1 gains over baseline smaller than those from prefills.

LLaVA follows similar patterns for D3~(2k) and GenImage~(2k) as illustrated in Appendix Fig.~\ref{fig:interval_progression_llava}.
The exception is DF40~(2k), where prefills show no confidence dip and underperform prompts.
Llama shows inconsistent patterns across datasets, with no clear relationship between confidence curves and performance (Appendix Fig.~\ref{fig:interval_progression_llama32}).

Overall, these results indicate that early overconfidence harms detection: baseline models select answers before examining the image closely, exhibiting high confidence despite limited analysis.
For some models, prefill-guided generation mitigates this effect by delaying answer selection and encouraging more deliberate reasoning.

\section{Discussion}

Our results show that prefilling responses for Vision-Language Models provides effective guidance for the detection of AI-generated images --- without task-specific fine-tuning.
We introduce a simple task-aligned prefill that consistently outperforms chain-of-thought across nearly all models and benchmarks, generalizing robustly across 16 state-of-the-art image generators --- even those producing photorealistic outputs lacking traditional artifacts.

We also find that responses must be prefilled directly, as instructing the model to begin its responses with the prefill does not reliably replicate the effect.
For two of the three models we tested, prefilling responses also outperforms standard user prompts.
As with prompts, careful phrasing is critical for effectiveness.

What might explain these improvements?
For Qwen and LLaVA on most datasets, prefills appear to counteract early overconfidence that otherwise leads to premature answers.
While baseline queries trigger immediate high confidence with poor performance, prefills induce uncertainty that persists throughout the model's response generation.
This delayed commitment leads to gradually improving detection performance --- suggesting that prefills may force more deliberate reasoning.

\section{Limitations}

A primary limitation of using VLMs for synthetic image detection is their computational cost.  
Compared to traditional models like CNNs or ViTs, Vision-Language Models require significantly more memory. 

While we found that Prefill-Guided Thinking improves synthetic image detection across three VLMs, we cannot assume it will generalize to all VLMs.
Architectures vary widely, as do their training regimes.
These models were also mid-sized (7--11B parameters), so effectiveness at larger scales remains untested.
Although we used state-of-the-art benchmarks from the past two years, the opaque nature of VLM training means we cannot exclude that the models were exposed to some of these images during pre-training.

While we identified a prefill suited to synthetic image detection, our finding that prefills are sensitive to phrasing means that the ease of discovering effective prefills for any given task remains an open question.

Our interpretability analysis showed overconfidence mitigation for Qwen and LLaVA on most datasets.
Llama, however, showed no clear relationship between confidence and performance.
Understanding why prefills affect some architectures but not others requires further study. 

\section*{Acknowledgments}

This work was supported by the Knight Foundation and the Luddy School of Informatics, Computing, and Engineering at Indiana University. 
We gratefully acknowledge NVIDIA for the GPU access that made this study possible. 
This work also used Jetstream2 at Indiana University through allocation CIS240194 from the Advanced Cyberinfrastructure Coordination Ecosystem.

\bibliography{references}

\pagebreak
\begin{onecolumn}
\appendix
\section{Appendix}
\label{sec:appendix}

\subsection{Full Reasoning Traces}
This section includes the full reasoning traces for all three methods. We can see the full traces produced by Qwen's \textbf{PGT} variations for the AI-generated image of a strawberry (Figs~\ref{fig:strawberryzeroshot}, ~\ref{fig:strawberryzeroshotcot}, ~\ref{fig:strawberryzeroshotss}).

\begin{figure*}[h] 
    \centering 
    \begin{minipage}[c]{0.5\textwidth} 
        
        \noindent 
        \begin{minipage}[t]{\linewidth} 
            \begin{usergrayprompt}
                User: [Image] Is this image real or AI‑generated?
            \end{usergrayprompt}
            \par\vspace{0.5mm} 
            \begin{assistantblueprompt}
                Assistant: The image appears to be real. The details, such as the texture of the strawberry's surface, the natural color gradient, and the lighting, are consistent with a photograph of a real strawberry. There are no visible artifacts or anomalies that would suggest it is AI-generated.
                \par\noindent\figureinput{Final Answer(real/ai‑generated):} real. \incorrectmark 
            \end{assistantblueprompt}
        \end{minipage}
        \vspace{4pt} 
    \end{minipage}
    \caption{An example of a baseline reasoning trace for detecting AI-generated images using a VLM (Qwen2.5-7B). Image from Fig.~\ref{fig:strawberry}.}
    \label{fig:strawberryzeroshot}
\end{figure*}

\begin{figure*}[h] 
    \centering 
    \begin{minipage}[c]{0.8\textwidth} 
        
        \noindent 
        \begin{minipage}[t]{\linewidth} 
            \begin{usergrayprompt}
                User: [Image] Is this image real or AI‑generated?
            \end{usergrayprompt}
            \par\vspace{0.5mm} 
            \begin{assistantblueprompt}
                Assistant \figureinput{Let's think step by step}:
                \par1. Texture and Detail: The strawberry appears to have a very realistic texture. The surface of the strawberry, including the tiny seeds and the green leafy top, looks natural and detailed.
                \par2. Lighting and Shadows: The lighting on the strawberry is well-balanced, with a soft shadow cast on the surface. This is typical of real-life photography where light interacts with the object in a natural way.
                \par3. Color and Contrast: The colors are vibrant and natural. The red of the strawberry and the green of the leafy top are consistent with what you would expect from a real strawberry.
                \par4. Overall Composition: The composition of the image is natural and unposed, which is more likely to be a real photograph rather than an AI-generated image.
                \par Given these observations, it is likely that this image is a real photograph of a strawberry. However, it's important to note that AI-generated images can sometimes be very realistic, so it's always a good idea to use multiple detection methods.
                \par\noindent\figureinput{Final Answer(real/ai‑generated):} real. \incorrectmark 
            \end{assistantblueprompt}
        \end{minipage}
        \vspace{4pt} 
    \end{minipage}
    \caption{An example of a chain-of-thought reasoning trace for detecting AI-generated images using a VLM (Qwen2.5-7B). Image from Fig.~\ref{fig:strawberry}.}
    \label{fig:strawberryzeroshotcot}
\end{figure*}

\begin{figure*}[h!] 
    \centering 
    \begin{minipage}[c]{0.8\textwidth} 
        
        \noindent 
        \begin{minipage}[t]{\linewidth} 
            \begin{usergrayprompt}
                User: [Image] Is this image real or AI‑generated?
            \end{usergrayprompt}
            \par\vspace{0.5mm} 
            \begin{assistantblueprompt}
                Assistant \figureinput{Let's examine the style and synthesis artifacts} in the image:
                \par1. Texture and Detail: The strawberry appears to have a very smooth and uniform texture, which is somewhat unusual for a real strawberry. Real strawberries typically have more varied and natural texture, with slight imperfections and variations in color and surface.
                \par2. Lighting and Reflection: The lighting on the strawberry is very even and lacks the natural variations you would expect in a real fruit. The reflections and highlights are too perfect and uniform, which is a common sign of AI-generated images.
                \par3. Color and Contrast: The color of the strawberry is vibrant and consistent, but the contrast is too high. Real strawberries have a more subtle contrast between the red and the green leaves, and the overall color gradient is not as sharp.
                \par4. Overall Composition: The strawberry is perfectly centered and the composition is very clean, which is often a characteristic of AI-generated images. Real photographs tend to have more natural imperfections and less perfect symmetry.
                \par Based on these observations, it is likely that this image is AI-generated.
                \par\noindent\figureinput{Final Answer(real/ai‑generated):} ai-generated. \correctmark 
            \end{assistantblueprompt}
        \end{minipage}
        \vspace{4pt} 
    \end{minipage}
    \caption{An example of a \textbf{S2} reasoning trace for detecting AI-generated images using a VLM (Qwen2.5-7B). Image from Fig.~\ref{fig:strawberry}.}
    \label{fig:strawberryzeroshotss}
\end{figure*}

\subsection{Detection Recall}
This section shows the detection recall for Qwen and LLaVA (Figs~\ref{fig:recall_radar_qwen}, ~\ref{fig:recall_radar_llava}).

\begin{figure*}[h]
\centering
\includegraphics[width=\textwidth]{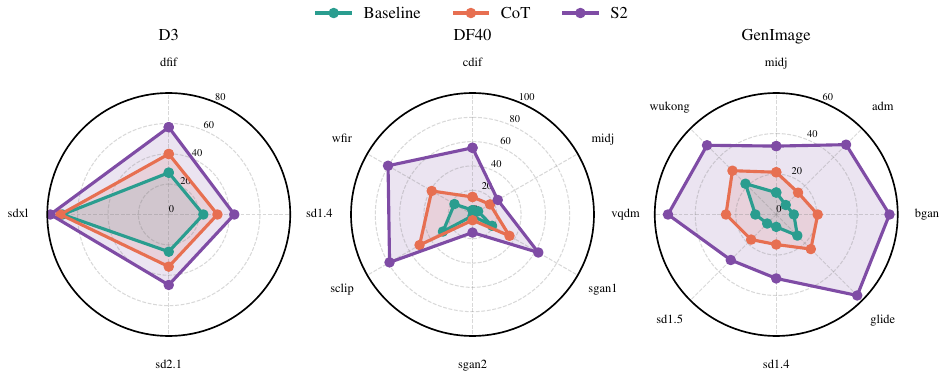}
\caption{
Detection recall (\%) for Qwen on each dataset, broken down by generator.
}
\label{fig:recall_radar_qwen}
\end{figure*}

\begin{figure*}[h]
\centering
\includegraphics[width=\textwidth]{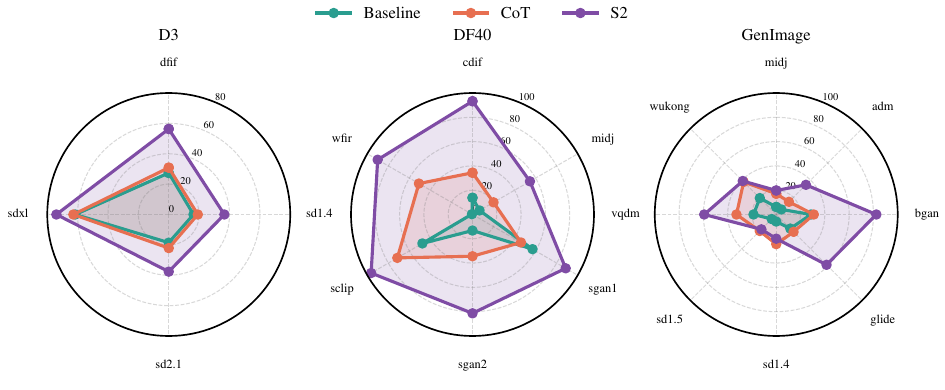}
\caption{
Detection recall (\%) for LLaVA on each dataset, broken down by generator.
}
\label{fig:recall_radar_llava}
\end{figure*}

\subsection{Guidance Comparison}
This section shows the detection performance for LLaVA and Llama across guidance variations of chain-of-thought and \textbf{S2} (Tables~\ref{tab:macro_f1_modes_llava_onevision_7b}, ~\ref{tab:macro_f1_modes_llama32_vision_11b}).

\begin{table*}[h]
    \centering
    \begin{tabular}{lllll}
    \toprule
    \textbf{Phrase} & \textbf{Type} & \textbf{D3~(2k)} & \textbf{DF40~(2k)} & \textbf{GenImage~(2k)} \\
    \midrule
    Baseline & --- & 41.8 & 49.3 & 45.1 \\
    \midrule
    \multirow{3}{*}{CoT} & Prefill & 44.2 & 59.2 & 55.1 \\
     & Pseudo-Prefill & 41.3 {\scriptsize (-2.9)} & 49.1 {\scriptsize (-10.1)} & 45.6 {\scriptsize (-9.5)} \\
     & Prompt & 43.5 {\scriptsize (-0.7)} & 52.2 {\scriptsize (-7.0)} & 55.4 {\scriptsize (+0.3)} \\
    \cmidrule(lr){1-5}
    \multirow{3}{*}{S2} & Prefill & 52.1 & 57.5 & 59.6 \\
     & Pseudo-Prefill & 41.6 {\scriptsize (-10.5)} & 49.2 {\scriptsize (-8.3)} & 45.6 {\scriptsize (-14.0)} \\
     & Prompt & 45.0 {\scriptsize (-7.1)} & 60.6 {\scriptsize (+3.1)} & 52.2 {\scriptsize (-7.4)} \\
    \bottomrule
    \end{tabular}
\caption{Macro F1 scores (\%) across datasets and guidance modes for LLaVA. Values in parentheses show differences from the corresponding prefill.}
\label{tab:macro_f1_modes_llava_onevision_7b}
\end{table*}

\begin{table*}[h]
    \centering
    \begin{tabular}{lllll}
    \toprule
    \textbf{Phrase} & \textbf{Type} & \textbf{D3~(2k)} & \textbf{DF40~(2k)} & \textbf{GenImage~(2k)} \\
    \midrule
    Baseline & --- & 47.7 & 40.1 & 60.4 \\
    \midrule
    \multirow{3}{*}{CoT} & Prefill & 47.8 & 60.3 & 58.1 \\
     & Pseudo-Prefill & 40.7 {\scriptsize (-7.1)} & 50.3 {\scriptsize (-10.0)} & 47.4 {\scriptsize (-10.7)} \\
     & Prompt & 50.5 {\scriptsize (+2.7)} & 61.5 {\scriptsize (+1.2)} & 62.9 {\scriptsize (+4.8)} \\
    \cmidrule(lr){1-5}
    \multirow{3}{*}{S2} & Prefill & 58.4 & 62.1 & 71.8 \\
     & Pseudo-Prefill & 58.6 {\scriptsize (+0.2)} & 58.7 {\scriptsize (-3.4)} & 64.4 {\scriptsize (-7.4)} \\
     & Prompt & 56.5 {\scriptsize (-1.9)} & 63.1 {\scriptsize (+1.0)} & 76.6 {\scriptsize (+4.8)} \\
    \bottomrule
    \end{tabular}
\caption{Macro F1 scores (\%) across datasets and guidance modes for Llama. Values in parentheses show differences from the corresponding prefill.}
\label{tab:macro_f1_modes_llama32_vision_11b}
\end{table*}

\subsection{Confidence Analysis}

This section expands the partial response analysis to LLaVA and Llama (Figs~\ref{fig:interval_progression_llava}, ~\ref{fig:interval_progression_llama32}).

\begin{figure*}[h]
\centering
\includegraphics[width=\textwidth]{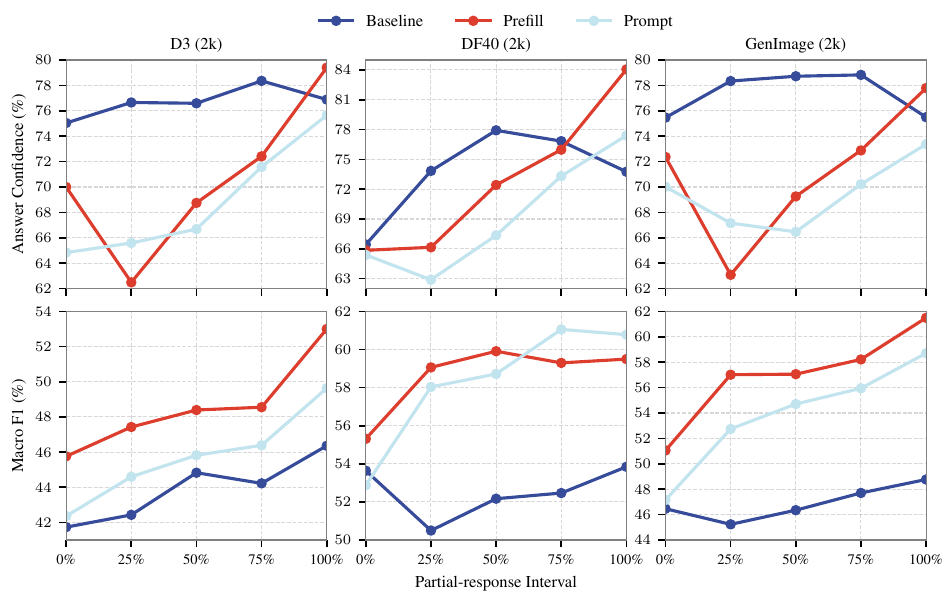}
\caption{Evolution of answer confidence and Macro F1 across partial responses for LLaVA on D3~(2k), DF40~(2k), and GenImage~(2k) using baseline, prefill, and prompt guidance modes.}
\label{fig:interval_progression_llava}
\end{figure*}

\begin{figure*}[h]
\centering
\includegraphics[width=\textwidth]{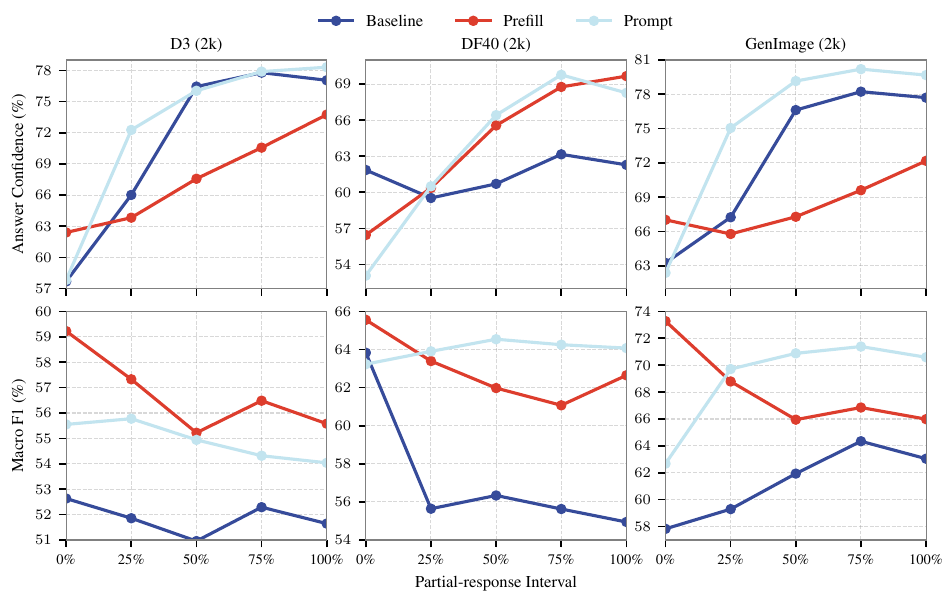}
\caption{Evolution of answer confidence and Macro F1 across partial responses for Llama on D3~(2k), DF40~(2k), and GenImage~(2k) using baseline, prefill, and prompt guidance modes.}
\label{fig:interval_progression_llama32}
\end{figure*}

\end{onecolumn}

\end{document}